\documentclass[sigconf]{acmart}

\AtBeginDocument{%
  }

\usepackage{algorithm}
\usepackage{algorithmic}
\usepackage{multirow}
\usepackage{colortbl}
\usepackage{enumitem}
\usepackage{pifont}
\setcopyright{acmlicensed}
\copyrightyear{2023}
\acmYear{2023}

\acmConference[MM '23] {Proceedings of the 31st ACM International Conference on Multimedia}{October 29--November 3, 2023}{Ottawa, ON, Canada.}

\acmBooktitle{Proceedings of the 31st ACM International Conference on Multimedia (MM '23), October 29--November 3, 2023, Ottawa, ON, Canada}

\acmPrice{15.00}
\acmDOI{10.1145/3581783.3612070}
\acmISBN{979-8-4007-0108-5/23/10}

\begin{document}

\title{Improving the Transferability of Adversarial Examples with Arbitrary Style Transfer}

\author{Zhijin Ge}
\email{zhijinge@stu.xidian.edu.cn}
\affiliation{%
  \institution{Xidian University}
  \city{}
  \country{}
}

\author{Fanhua Shang}
\authornote{Corresponding author}
\email{fhshang@tju.edu.cn}
\affiliation{%
  \institution{Tianjin University}
  \city{}
  \country{}
}

\author{Hongying Liu}
\authornotemark[1]
\email{hyliu2009@tju.edu.cn}
\affiliation{%
  \institution{Tianjin University}
  \city{}
  \country{}
}

\author{Yuanyuan Liu}
\authornotemark[1]
\email{yyliu@xidian.edu.cn}
\affiliation{%
 \institution{Xidian University}
 \city{}
 \country{}
}

\author{Liang Wan}
\email{lwan@tju.edu.cn}
\affiliation{%
  \institution{Tianjin University}
  \city{}
  \country{}
}

\author{Wei Feng}
\email{wfeng@tju.edu.cn}
\affiliation{%
  \institution{Tianjin University}
  \city{}
  \country{}
}

\author{Xiaosen Wang}
\email{xiaosen@hust.edu.cn}
\affiliation{%
  \institution{Huawei Singular Security Lab}
  \city{}
  \country{}
}

\renewcommand{\shortauthors}{Zhijin Ge et al.}
\begin{abstract}
  Deep neural networks are vulnerable to adversarial examples crafted by applying human-imperceptible perturbations on clean inputs. Although many attack methods can achieve high success rates in the white-box setting, they also exhibit weak transferability in the black-box setting. Recently, various methods have been proposed to improve adversarial transferability, in which the input transformation is one of the most effective methods. In this work, we notice that existing input transformation-based works mainly adopt the transformed data in the same domain for augmentation. Inspired by domain generalization, we aim to further improve the transferability using the data augmented from different domains. Specifically, a style transfer network can alter the distribution of low-level visual features in an image while preserving semantic content for humans. Hence, we propose a novel attack method named Style Transfer Method (STM) that utilizes a proposed arbitrary style transfer network to transform the images into different domains. To avoid inconsistent semantic information of stylized images for the classification network, we fine-tune the style transfer network and mix up the generated images added by random noise with the original images to maintain semantic consistency and boost input diversity. Extensive experimental results on the ImageNet-compatible dataset show that our proposed method can significantly improve the adversarial transferability on either normally trained models or adversarially trained models than state-of-the-art input transformation-based attacks. Code is available at: https://github.com/Zhijin-Ge/STM.
\end{abstract}



\maketitle

\section{Introduction}
\label{sec:intro}

Deep neural networks (DNNs) have achieved great performance in various computer vision tasks, such as image classification \cite{AlexKrizhevsky2012ImageNetCW, KarenSimonyan2014VeryDC}, face recognition \cite{MahmoodSharif2016AccessorizeTA}, and autonomous driving \cite{AnaIMaqueda2018EventbasedVM, WeibinWu2019DeepVT}. However, numerous works have shown that DNNs are vulnerable to adversarial examples \cite{ChristianSzegedy2013IntriguingPO, SeyedMohsenMoosaviDezfooli2015DeepFoolAS, IanGoodfellow2014ExplainingAH, YinpengDong2017BoostingAA, wang2019gan, YueLi021,zhang2022practical}, in which applying human-imperceptible perturbations on clean input can result in misclassification. Moreover, the adversaries often exhibit transferability across neural network models, \textit{i.e.}, the adversarial examples generated on one model might also mislead other models \cite{YinpengDong2017BoostingAA, JiadongLin2019NesterovAG, YinpengDong2019EvadingDT, CihangXie2018ImprovingTO}. Such transferability brings a great threat to crucial security applications since it is possible for hackers to attack a real-world DNN application without knowing any information about the target model. 
Thus, learning the transferability of adversarial examples can help us understand the vulnerabilities of neural network applications and improve their robustness against adversaries in the real world.

\begin{figure}[t]
  \centering
  \includegraphics[width=0.48 \textwidth]{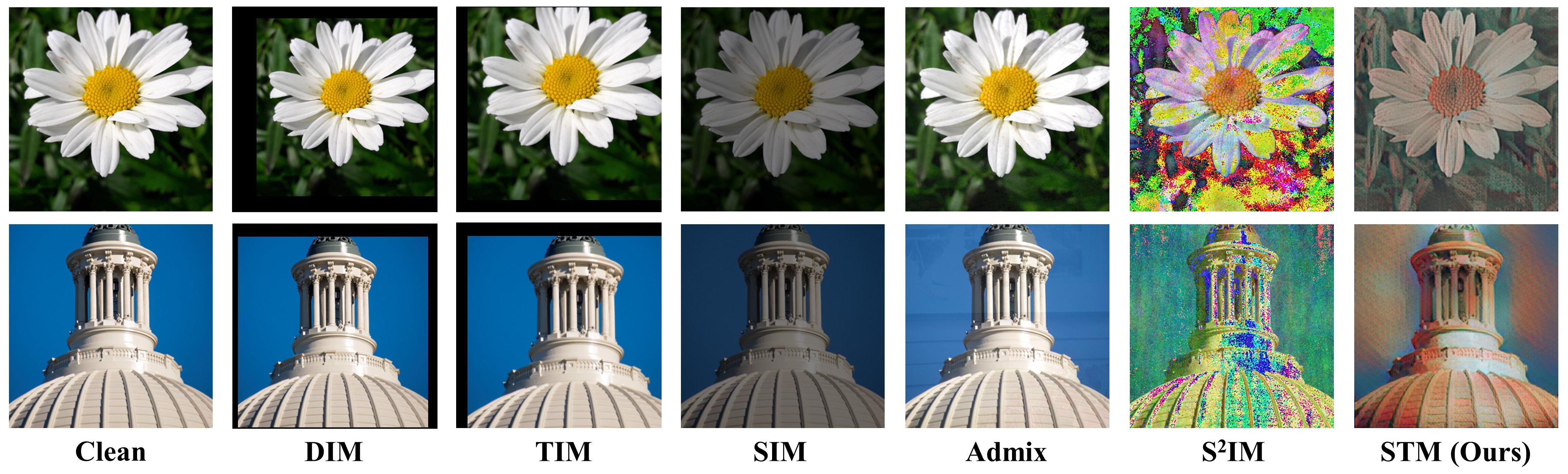}
  \caption{
  Illustration of two transformed images by various input transformation-based attacks. Our STM generates some images in different domains by using our style augmentation module, leading to different styles from clean images but maintaining the semantic contents compared with existing transformation-based attacks.
  }
  \label{fig:1}
\end{figure}

Although several attack methods \cite{IanGoodfellow2014ExplainingAH, AlexeyKurakin2016AdversarialEI, carlini2017towards} have exhibited great attack performance in the white-box setting, they have low transferability when attacking black-box models, especially for some advanced defense models \cite{AleksanderMadry2018TowardsDL, FlorianTramr2017EnsembleAT}. To improve the transferability of the adversarial examples, various approaches have been proposed, such as introducing momentum into iterative gradient-based attacks for optimization \cite{YinpengDong2017BoostingAA, JiadongLin2019NesterovAG,zhang2023improving}, utilizing an ensemble of source models \cite{YanpeiLiu2016DelvingIT}, and input transformations \cite{JiadongLin2019NesterovAG, YinpengDong2019EvadingDT, CihangXie2018ImprovingTO, XiaosenWang2021AdmixET, YuyangLong2022FrequencyDM,wang2023structure}. Among these strategies, the input transformation-based attack method is one of the most effective approaches to enhance the transferability using data transformations, such as randomly resizing and padding, translating, scaling, and mixing up different images. By analysis these methods, we notice that they mainly adopt the transformed data in the \textit{same source domain} for augmentation, which might limit the adversarial transferability. 

Lin \textit{et al.} \cite{JiadongLin2019NesterovAG} treated the process of generating adversarial examples on the white-box model as a standard neural network training process, in which the adversarial transferability is equivalent to the model generalization. As we all know, \textit{domain bias} \cite{lin2022bayesian} decays the model generalization, \textit{i.e.}, the model trained on a specific data domain cannot generalize well to other domain datasets. Similarly, there is domain bias between different models due to various architectures and randomness during the training. Recently, several studies \cite{carlucci2019domain, zhou2020deep} address the \textit{domain bias} issue by training the models using data from different domains to reduce the risk of overfitting to the source domain and improve the model generalization ability. With the analogy between adversarial transferability and model generalization, it inspires us to utilize the data from different domains to improve adversarial transferability.

In practice, it is usually expensive to obtain data from different domains, let alone images from different domains with the same semantic label. Thanks to the success of DNNs, style transfer has made great progress, which can alter the distribution of low-level visual features in an image whilst preserving semantic contents for humans \cite{LeonAGatys2016ImageST, PhilipTJackson2019StyleAD}. Recently, image style transfer acted as an effective data augmentation technique to boost the generalization across different domains \cite{PhilipTJackson2019StyleAD, KaiyangZhou2021DomainGW}. This inspires us to transform the data using a style transfer network and propose a new Style Transfer Method (STM) to boost adversarial transferability.

Specifically, we introduce a new arbitrary style transfer network to transfer the style of a clean input image, which introduces data from different domains for augmentation. Since the stylized images may mislead the surrogate model, this can result in imprecise gradients during the iterative optimization process. To address this issue, we first fine-tune our style transfer network so that the generated images can be correctly classified by multiple models. We also mix up the original image with its style-transformed images to further avoid such imprecise gradients. For a more stable gradient update, we adopt the average gradient of multiple transformed images with random noise to update the perturbation.
We illustrate some transformed images by various input transformation-based attacks in Figure \ref{fig:1}. As we can see, there is no significant visual difference between the clean images and transformed images by DIM \cite{CihangXie2018ImprovingTO}, TIM \cite{YinpengDong2019EvadingDT}, SIM \cite{JiadongLin2019NesterovAG} and Admix \cite{XiaosenWang2021AdmixET}. As S$^2$IM \cite{YuyangLong2022FrequencyDM} transforms the image in the frequency domain, it significantly changes the images and introduces semantic shift. Conversely, STM changes the style but maintains the semantic content, which changes the low-level statistical features (\textit{e.g.}, texture, contrast) of the clean image and makes the transformed image deviate from the source domain. It significantly enhances the diversity of the input images for gradient calculation and results in better transferability. We summarize our contributions as follows:

\begin{itemize}[leftmargin=*,noitemsep,topsep=2pt]
	\item We find that existing input transformation-based methods mainly adopt the transformed data in the same domain, which might limit the adversarial transferability. To address this limitation, we propose a novel attack method, which introduces data from different domains to enhance the adversarial transferability. 
	\item We propose a new input transformation by mixing up the input images with the transformed image generated by our fine-tuned style transfer network and adding random noise for diverse images from various domains.
	\item Empirical evaluations show that our STM can significantly boost transferability on either normally or adversarially trained models. In particular, STM outperforms state-of-the-art methods by a margin of 7.45\% on average for adversarially trained models.
\end{itemize}

\section{Related Work}
\label{sec:relatedwork}
This section provides a brief overview of the adversarial attacks and the improved domain generalization with style transfer networks.

\subsection{Adversarial Attacks}
After Szegedy \textit{et al.} \cite{ChristianSzegedy2013IntriguingPO} identified adversarial examples, many attacks have been developed to generate adversarial examples. Goodfellow \textit{et al.} \cite{IanGoodfellow2014ExplainingAH} proposed the Fast Gradient Sign Method (FGSM) to generate adversarial examples with one step of gradient update. Kurakin \textit{et al.} \cite{AlexeyKurakin2016AdversarialEI} further extend FGSM to an iterative version with a smaller step size $\alpha$, denoted as I-FGSM, which exhibits superior attack success rates in the white-box setting. On the other hand, black-box attacks are more practical since they only access limited or no information about the target model. Query-based attacks \cite{WielandBrendel2017DecisionBasedAA, PinYuChen2017ZOOZO, wang2022triangle} often take hundreds or even thousands of queries to generate adversarial examples, making it inefficient. By contrast, transfer-based attacks \cite{YanpeiLiu2016DelvingIT, YinpengDong2017BoostingAA,wang2023rethinking} generate the adversarial on the surrogate model without accessing the target model, leading to great practical applicability and attracting increasing attention.

Unfortunately, although I-FGSM has exhibited great effectiveness in the white-box setting, it has low transferability when attacking black-box models. To boost the adversarial transferability,  Liu \textit{et al.} \cite{YanpeiLiu2016DelvingIT} proposed an ensemble-model attack that attacks multiple models simultaneously. Dong \textit{et al.} \cite{YinpengDong2017BoostingAA} integrated momentum into I-FGSM (called MI-FGSM) to stabilize the update direction. Lin \textit{et al.} \cite{JiadongLin2019NesterovAG} adopted Nesterov's accelerated gradient to further enhance the transferability. Wang \textit{et al.} \cite{XiaosenWang2021EnhancingTT} considered the gradient variance of the previous iteration to tune the current gradient. Wang \textit{et al.} \cite{XiaosenWang2021BoostingAT} proposed enhanced momentum by accumulating the gradient of several data points in the direction of the previous gradient for better transferability.

Inspired by the data augmentation strategies~\cite{chlap2021review, nanni2020data}, various input transformation methods have been proposed to effectively boost adversarial transferability. Xie \textit{et al.} \cite{CihangXie2018ImprovingTO} proposed to adopt diverse input patterns by randomly resizing and padding to generate transferable adversarial examples. Dong \textit{et al.} \cite{YinpengDong2019EvadingDT} used a set of translated images to optimize the adversarial perturbations, and approximated such process by convolving the gradient at untranslated images with a kernel matrix for high efficiency. Lin \textit{et al.} \cite{JiadongLin2019NesterovAG} leveraged the scale-invariant property of DNNs and thus averaged the gradients \textit{w.r.t.} different scaled images to update adversarial examples. Wang \textit{et al.} \cite{XiaosenWang2021AdmixET} mixed up a set of images randomly sampled from other categories while maintaining the original label of the input to craft more transferable adversaries. Long \textit{et al.} \cite{YuyangLong2022FrequencyDM} proposed a novel spectrum simulation attack to craft more transferable adversarial examples by transforming the input image in the frequency domain.

\subsection{Domain Generalization with Style Transfer}
The success of DNNs heavily relies on the \textit{i.i.d.} assumption, \textit{i.e.}, training and testing datasets are drawn from the same distribution. When such an assumption is violated, DNNs usually suffer from severe performance degradation \cite{BenjaminRecht2019DoIC}. A typical solution to \textit{domain bias} is transfer learning, in which a network is pre-trained on a related task with a large dataset and then fine-tuned on a new dataset \cite{shao2014transfer, yosinski2014transferable}. However, transfer learning needs to reuse the same architecture as that of the pre-trained network and careful application of layer freezing to prevent the prior knowledge from being forgotten during fine-tuning. Domain adaptation is another way to address domain bias, which encompasses a variety of techniques for adapting a model post-training to improve its accuracy on a specific target domain. It is often implemented by minimizing the distance between the source and the target feature domains in some manner \cite{hoffman2017simultaneous, li2016revisiting}. Though domain adaptation is usually effective, its functionality is limited since it can only help a model generalize to a specific target domain. Domain generalization \cite{carlucci2019domain, zhou2020deep} aims to address the issue by learning data from diverse domains to boost the model generalization on other unseen domains. Recently, various style augmentation methods have been proposed to explore domain generalization. Wang \textit{et al.} \cite{wang2021learning} proposed a style-complement module that generates stylized images to enhance the generalization on unseen target domains. Jackson \textit{et al.} \cite{PhilipTJackson2019StyleAD} proposed data augmentation-based random style transfer to improve the robustness of the convolutional neural networks, which can also enhance the robustness to domain shift. Inspired by the above works, we postulate that introducing data from different domains to craft adversarial examples can also improve adversarial transferability.

\section{Methodology}
\label{sec:method}
In this section, we first provide details of several input transformation-based attacks. Then we explain our motivation and propose a new attack method named Style Transfer Method (STM) to generate adversarial examples. Finally, we will discuss the differences with other input transformation-based methods.

\begin{figure*}[t]
	\centering\includegraphics[width=0.85 \textwidth]{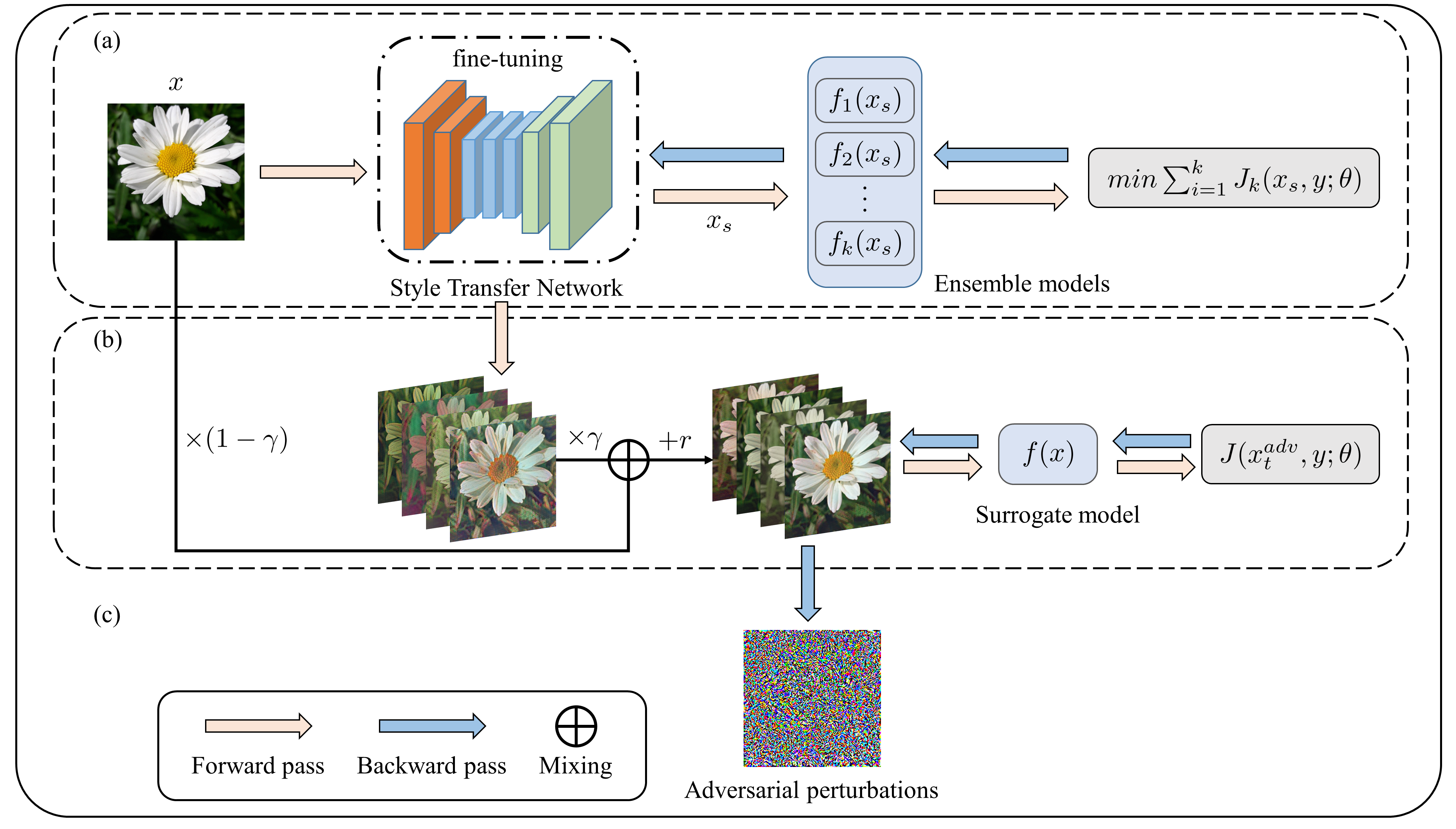}
	\caption{The overall framework of our proposed style transfer attack method. (a): We fine-tune the pre-trained style transfer network such that the generated stylized images keep the original semantic labels as much as possible for the classification networks. (b): Combine the original image with the generated stylized images by the mixing ratio $\gamma$ to retain semantic consistency, and add random noise for image augmentation. (c): Add the generated perturbations to the original image to generate the adversarial example.}
	\label{fig:2}
\end{figure*}

\subsection{Preliminaries}
  \textbf{Notation.} Given a classifier $f(\mathbf{x}; \theta)$ with parameters $\theta$, and a benign input image $\mathbf{x}$ with ground-truth label $y$. Let $J(\mathbf{x}, y; \theta)$ be the loss function (\textit{e.g.}, the cross-entropy loss). Specifically, there are two categories of adversarial attacks, \textit{i.e.}, non-targeted and targeted attacks. Non-targeted attack searches an adversarial example $\mathbf{x}^{adv}$ that satisfies $\|\mathbf{x}-\mathbf{x}^{adv} \| _{p}\le \epsilon $ but misleads the classifier ($f(\mathbf{x}^{adv};\theta)\ne y$). Targeted attack fools the classifier into outputting a specific label ($f(\mathbf{x}^{adv};\theta)= y^*$). Here $\epsilon$ is the maximum magnitude of perturbation, $y^*$ is the target label, and $p$ could be $0,2,$ or $\infty$. To align with other works, we set $p=\infty$ in this work.

Input transformation-based attacks are typical methods for boosting the adversarial transferability, \textit{e.g.}, Diverse Input Method (DIM) \cite{CihangXie2018ImprovingTO}, Translation-Invariant Method (TIM) \cite{YinpengDong2019EvadingDT} and Scale-Invariant Method (SIM) \cite{JiadongLin2019NesterovAG}. Besides these methods, several more powerful methods such as Admix \cite{XiaosenWang2021AdmixET} and S$^2$IM \cite{YuyangLong2022FrequencyDM} were proposed, and we will introduce these two methods in this subsection in detail.

\textbf{Admix Attack Method.}
Admix \cite{XiaosenWang2021AdmixET} proposes to mix a set of images randomly sampled from other categories while retaining the original label of the input to craft more transferable adversaries. This method can integrate with the SIM method, and its average gradient can be expressed as follows:
\begin{equation}
  \bar{\mathbf{g}}_{t}=\frac{1}{m_{1} \cdot m_{2}} \sum_{\mathbf{x}^{\prime} \in \mathbf{X}^{\prime}} \sum_{i=0}^{m_{1}-1} \nabla_{\mathbf{x}_t^{adv}} J\left(\left(\mathbf{x}_{t}^{adv}+\eta \cdot \mathbf{x}^{\prime}\right)/2^i, y ; \theta\right),
\end{equation}
where $\eta$ controls the strength of admixed images, $m_1$ is the number of admixed images for each $\mathbf{x}^{\prime}$, and $\mathbf{X}^{\prime}$ denotes the set of $m_2$ randomly sampled images from other categories.

\textbf{S$^2$IM Attack Method.}
S$^2$IM \cite{YuyangLong2022FrequencyDM} applies a spectrum transformation to input data, thus performs the model augmentation in the frequency domain, and proposes the transformation based on the discrete cosine transform ($\mathcal{D}$) and inverse discrete cosine transform ($\mathcal{D}_{\mathcal{I}}$) techniques to diversify input images.
\begin{equation}
	\bar{\mathbf{g}}_t = \nabla_{\mathbf{x}_t^{adv}} J(\mathcal{T}(\mathbf{x}_t^{adv}),y;\theta),
\end{equation}
where $\mathcal{T}(\mathbf{x}) = \mathcal{D}_{\mathcal{I}}(\mathcal{D}(\mathbf{x}+\xi) \odot \mathbf{M})$, $\odot$ is the Hadamard product, $\xi \sim \mathcal{N}(0, \sigma^2)$ and each element of $\mathbf{M} \sim \mathcal{U}(1-\rho, 1+\rho)$ is random variant sampled from Gaussian and uniform distribution.

Generally, the above input transformation-based methods are often integrated into MI-FGSM \cite{YinpengDong2017BoostingAA}.
\begin{equation}
	\mathbf{g}_{t+1}=\mu \cdot \mathbf{g}_t+\frac{\bar{\mathbf{g}}_t}{\left \| \bar{\mathbf{g}}_t \right \|_1 },
\end{equation}
where $\mathbf{g}_0=0$, and the adversarial examples can be generated by $\mathbf{x}_{t+1}^{adv}=\mathbf{x}_{t}^{adv}+\epsilon \cdot \mbox{sign}(\mathbf{g}_{t+1})$.

S$^2$IM performs data augmentation in the frequency domain, which inspires us to explore the data from different domains to improve the attack transferability. Admix mixes images from other categories to obtain diverse inputs and also motivates us to retain the semantic labels of stylized images by mixing up the original image content. The differences between our method and these two methods will be shown in Section \ref{sec:difference}.  
\subsection{Motivation}
Lin \textit{et al.} \cite{JiadongLin2019NesterovAG} analogized the generation of adversarial examples with standard neural network training and considered that the transferability of adversarial examples is related to the generalization of normally trained models. Therefore, some existing methods to improve attack transferability are mainly from the perspective of optimization \cite{YinpengDong2017BoostingAA, JiadongLin2019NesterovAG, XiaosenWang2021EnhancingTT} or data augmentation \cite{XiaosenWang2021AdmixET, JiadongLin2019NesterovAG, YuyangLong2022FrequencyDM}. 

In this paper, we notice that the \textit{domain bias} issue also affects the generalization ability of the normally trained model. For example, a model trained on a specific data domain cannot generalize well to other domain datasets. Besides, even the same architecture network trained on two datasets from different domains will result in two different model parameters, which causes a bias issue between models. In the black-box setting, this domain bias issue also exists due to the structural differences between black-box models and the source model, which limits the transferability of the adversarial examples.

In the domain generalization fields, several studies \cite{carlucci2019domain, zhou2020deep} address the \textit{domain bias} issue by training the models using data from different domains to improve the model generalization ability. On the contrary, we find that previous input transformation-based methods mainly apply data augmentation in the same source domain, which might limit the adversarial transferability. It inspires us to utilize data from different domains to improve adversarial transferability. However, data from different domains are often expensive and difficult to obtain. Thanks to the development of style transfer models and domain generalization with style transfer studies \cite{PhilipTJackson2019StyleAD, XunHuang2017ArbitraryST, VincentDumoulin2016ALR, KarenSimonyan2015VeryDC, DmitryUlyanov2016TextureNF}, we can quickly obtain a large amount of data from different domains. 

Based on the above analysis, we propose a new attack method named Style Transfer Method (STM) to boost adversarial transferability, which transforms the data into different domains by using an arbitrary style transfer network. It is worth noting that we are not directly using the stylized image for the attack, we mainly use the gradient information of the stylized images during the iterations to obtain more effective adversarial perturbations.

\subsection{Style Transfer Method}
The overall framework of our Style Transfer Method (STM) is shown in Figure \ref{fig:2}, which we will cover in detail below.

\textbf{Arbitrary style transfer.}
An important component of our method is the style transfer network $ST(\cdot)$, which can replace the style of an input image $\mathbf{x}$ with that of an arbitrary style image $\mathbf{s}$. To apply a specific style, the network must observe the chosen style image. This is accomplished through a style predictor network $P(\cdot)$, which maps a style image $\mathbf{s}$ to a style embedding $\mathbf{z}=P(\mathbf{s})$, where $\mathbf{z}$ is a feature vector of the $\mathbf{s}$. The style embedding influences the action of the style transfer network via conditional instance normalization \cite{VincentDumoulin2016ALR}, and the style transfer network $ST(\mathbf{x}, P(\mathbf{s}))$, as a typical encoder/decoder architecture, generates the corresponding style images specifically for the normalized parameters of each style.

In general, the style predictor network predicts the feature maps of a style image $\mathbf{s}$ and embeds them into the style transfer network to generate a specific style of image. Jackson \textit{et al.} \cite{PhilipTJackson2019StyleAD} suggested that rather than providing randomly selected style images through a style predictor to generate random style embeddings, it would be more computationally efficient to simulate this process by sampling them directly from a probability distribution, which was trained on about 79,433 artistic images. From this strategy, the arbitrary style images can be obtained as follows:
\begin{equation}
	\mathbf{x}_s = ST(\mathbf{x}, \mathbf{z}), \;\;  \mathbf{z} = (1-\upsilon ) \cdot \mathcal{N}(\mathbf{\mu}_s, \mathbf{\Sigma})+\upsilon \cdot P(\mathbf{x}),
	\label{eq:stylex}
\end{equation}
where $\upsilon$ is the style embedding interpolation parameter, $\mathbf{\mu}_s$ and $\mathbf{\Sigma}$ are the empirical mean and covariance matrix of the style image embeddings $P(\mathbf{s})$. Here, $\mathbf{\mu}_s=\mathbb{E}_s[P(\mathbf{s})], \; \mathbf{\Sigma}_{i,j}=Cov [P(\mathbf{s})_i, P(\mathbf{s})_j]$.

In this work, we simplify the process of image style transfer to adapt the process of generating adversarial examples. Specifically, the style predictor network is not necessary for us due to the fact that our inputs are clean images, and to obtain an arbitrary style image, we replace the style embedding vector $\mathbf{z}$ with a standard orthogonal distribution vector. Thus, the arbitrary style images of the adversarial examples at the $t-$th iteration can be obtained as:
\begin{equation}
	\mathbf{x}_s = ST(\mathbf{x}_t^{adv}, \mathbf{z}), \;\;  \mathbf{z} = \mathcal{N}(0, 1).
    \label{eq:stylex}
\end{equation}
It is an available method since a standard orthogonal distribution can be transformed by a series of affine transformations to obtain an arbitrary style embedding vector \cite{VincentDumoulin2016ALR, XunHuang2017ArbitraryST}.

\textbf{Preserving semantic consistency of stylized images.}
Although style transfer networks can generate stylized images while preserving semantic content for humans, they change low-level features of original images (\textit{e.g.}, texture, color, and contrast), which might change the semantic labels of stylized images. Since the stylized images may mislead the surrogate model, this will lead to imprecise gradient information during iterations and affect the success rate of adversarial attacks.
 
To address this issue, as shown in Figure \ref{fig:2} (a), we first fine-tune our style transfer network by using a classifier that integrated several classification models. Specifically, we expect that the images generated by the style transfer network can still be correctly classified by the classifier. We define the fine-tuning loss function as follows:
\begin{equation}
	\label{eq:fine_tune_loss}
	L(\mathbf{x})=\sum_{i=1}^{k}w_kJ_k(ST(\mathbf{x}, \mathbf{z}), y; \theta),
\end{equation}
where $J_k(\cdot)$ is the $k-$th model cross-entropy loss, and $\sum_{i=1}^{k}w_k=1$ is the ensemble weight. For each input image, we keep the parameter weight of the classification model constant and use gradient descent to minimize the loss function $L(\mathbf{x})$ to update the model parameters of the style transfer network $ST(\cdot)$ during the model fine-tuning process. In detail, we integrate Inception-v3, Inception-v4 \cite{ChristianSzegedy2016RethinkingTI}, ResNet-101, and ResNet-152 \cite{KaimingHe2015DeepRL} classification models and average the cross-entropy loss of these models. We randomly select $1,000$ images on the Imagenet-$1000$ \cite{AlexKrizhevsky2012ImageNetCW} validation dataset as our fine-tuning dataset, and each of these images can be correctly classified by neural networks. Then we fine-tune the style augmentation module on them using the Adam optimizer, and we set 30 epochs with a learning rate of $1\times10^{-4}$.

Although fine-tuning the style transfer network can enable partial data recovery to the original semantic labels, the recognition accuracy of classifier networks for data from different domains is still limited due to the influence of \textit{domain bias}. Inconsistent semantic labels may generate imprecise gradient directions during gradient calculation, thus limiting the transferability of the adversarial examples. To address this limitation, we mix up the generated stylized images with the original image, as shown in Figure \ref{fig:2} (b), which can be expressed as follows: 

\begin{equation}
	\label{eq:miximg}
	\bar{\mathbf{x}} = \gamma \cdot \mathbf{x} + (1-\gamma) \cdot \mathbf{x}_s,
\end{equation}
where $\gamma$ is a mixing ratio, and $\gamma \in \left [ 0,1 \right ] $. 
Mixing up the original image with its style-transformed images allows the augmented image to introduce features from different domains while preserving the original semantic labels to avoid generating imprecise gradient information during the iterative optimization process. Lastly, we add stochastic noise $\mathbf{r}$ on the stylized images to obtain diverse images from different domains to enhance the transferability of the adversarial examples, where $\mathbf{r} \sim \mathcal{U}[-(\beta \cdot \epsilon)^d,(\beta \cdot \epsilon)^d]$, and $\beta$ is a given parameter.

After the above analysis, we propose a novel style transfer-based attack method to improve the attack transferability, and the proposed algorithm is outlined in Algorithm \ref{alg: STM}. 

\begin{algorithm}[t] 
	\caption{Style Transfer attack Method (STM)}
	\label{alg: STM}
	\begin{flushleft}
	\textbf{Input}: A clean image $\mathbf{x}$ with ground-truth label $y$, surrogate classifier with parameters $\theta$, and the loss function $J$.
	\end{flushleft}
	\begin{flushleft}
	\textbf{Parameters}: The magnitude of perturbation $\epsilon$; maximum iteration $T$; decay factor $\mu$; the upper bound of neighborhood $\beta$ for $\mathbf{r}$; the mixing ratio $\gamma$; the number of random generating examples $N$.
	\end{flushleft}
	\begin{flushleft}
	\textbf{Output}: An adversarial example $\mathbf{x}^{adv}$.
	\end{flushleft}
	
	\begin{algorithmic}[1]
		\STATE $\alpha = \epsilon/T$;
		\STATE $\mathbf{g}_0 = 0, $ $\mathbf{x}_0^{adv}=\mathbf{x}$;
		\FOR {$t = 0, 1, \cdots, T-1$}
		\FOR {$i=0, 1, \cdots, N-1$}
		\STATE Obtain a random stylized image $\mathbf{x}_s$ by $ST(\mathbf{x}_t^{adv}, \mathbf{z})$;
		
		\STATE Mix the original image by $\bar{\mathbf{x}} = \gamma \cdot \mathbf{x} + (1-\gamma) \cdot \mathbf{x}_s+\mathbf{r};$
		\STATE Calculate the gradient $\bar{\mathbf{g}}_i = \nabla_{\bar{\mathbf{x}}} J(\bar{\mathbf{x}}, y; \theta)$;
		\ENDFOR
		\STATE Get the average gradient, $\bar{\mathbf{g}}=\frac{1}{N} \sum\limits_{i=1}^{N} \bar{\mathbf{g}}_i$;
		\STATE $\mathbf{g}_{t+1}=\mu \cdot \mathbf{g}_t +\frac{\bar{\mathbf{g}}}{\left \| \bar{\mathbf{g}} \right \|_1 }$;
		\STATE Update $\mathbf{x}_{t+1}^{adv}$ by $$\mathbf{x}_{t+1}^{adv} =\mbox{Clip}_\mathbf{x}^{\epsilon} \{\mathbf{x}_t^{adv}+\alpha \cdot \mbox{sign}(\mathbf{g}_{t+1})\};$$
		\ENDFOR
		
		\STATE \textbf{return} $\mathbf{x}^{adv}=\mathbf{x}_T^{adv}$.
	\end{algorithmic}
\end{algorithm}

\begin{table*}[t]
      \centering
      \caption{Untargeted attack success rates (\%) of various input transformation-based attacks in the single model setting. The adversarial examples are crafted on Inc-v3, Inc-v4, IncRes-v2, and Res-101 by DIM, TIM, SIM, Admix, S$^2$IM, and our STM attack methods, respectively.  * indicates the white-box model.}
      \vspace{-1mm}
      \setlength{\tabcolsep}{9.50pt}
         \begin{tabular}{|c|c|ccccccc|c|}
         \hline
         Model & Attack & Inc-v3 & Inc-v4 & IncRes-v2 & Res-101 & Inc-v3$_{ens3}$\! & \! Inc-v3$_{ens4}$\! & \!\! IncRes-v2$_{ens}$\!  & Avg. \\
         \hline
         \hline
         \multirow{6}[2]{*}{Inc-v3} & DIM   & 99.7*  & 72.4  & 66.7  & 62.8  & 32.0    & 30.7  & 16.6  & 54.41 \\
         & TIM   & \textbf{100.0*} & 51.0    & 46.8  & 47.8  & 30.0    & 30.9  & 21.5  & 46.85 \\
         & SIM   & \textbf{100.0*} & 69.7  & 68.2  & 63.8  & 37.8  & 37.9  & 21.8  & 57.03 \\
         & Admix & \textbf{100.0*} & 78.6  & 75.1  & 69.5  & 40.9  & 41.7  & 23.1  & 61.27 \\
         & S$^2$IM   & 99.7  & 87.5  & 86.7  & 77.7  & 58.2  & 56.2  & 34.9  & 71.55 \\
         & \cellcolor[rgb]{0.9, 0.9, 0.9} \textbf{STM}  & \cellcolor[rgb]{0.9, 0.9, 0.9} 99.9*  & \cellcolor[rgb]{0.9, 0.9, 0.9} \textbf{90.8} & \cellcolor[rgb]{0.9, 0.9, 0.9} \textbf{90.1} & \cellcolor[rgb]{0.9, 0.9, 0.9} \textbf{82.4} & \cellcolor[rgb]{0.9, 0.9, 0.9} \textbf{68.3} & \cellcolor[rgb]{0.9, 0.9, 0.9} \textbf{68.1} & \cellcolor[rgb]{0.9, 0.9, 0.9} \textbf{46.3} & \cellcolor[rgb]{0.9, 0.9, 0.9} \textbf{77.98} \\
         \hline
         \multirow{6}[2]{*}{Inc-v4} & DIM   & 75.0    & 99.2*  & 68.8  & 71.7  & 29.0    & 26.2  & 16.6  & 55.21 \\
         & TIM   & 58.7  & 99.8*  & 47.7  & 58.9  & 28.0    & 28.2  & 20.8  & 48.87 \\
         & SIM   & 82.4  & \textbf{99.9*} & 74.0    & 80.7  & 46.6  & 44.2  & 30.8  & 65.51 \\
         & Admix & 85.7  & \textbf{99.9*} & 76.5  & 81.6  & 47.7  & 44.8  & 29.3  & 66.50 \\
         & S$^2$IM   & \textbf{90.4} & 99.6*  & 86.3  & 85.9  & 58.5  & 55.4  & 37.2  & 73.33 \\
         & \cellcolor[rgb]{0.9, 0.9, 0.9} \textbf{STM}   & \cellcolor[rgb]{0.9, 0.9, 0.9} 90.0    & \cellcolor[rgb]{0.9, 0.9, 0.9} 99.0*    & \cellcolor[rgb]{0.9, 0.9, 0.9} \textbf{86.4} & \cellcolor[rgb]{0.9, 0.9, 0.9} \textbf{86.1} & \cellcolor[rgb]{0.9, 0.9, 0.9} \textbf{61.0} & \cellcolor[rgb]{0.9, 0.9, 0.9} \textbf{59.2} & \cellcolor[rgb]{0.9, 0.9, 0.9} \textbf{41.2} & \cellcolor[rgb]{0.9, 0.9, 0.9} \textbf{74.70} \\
         \hline
         \multirow{6}[2]{*}{IncRes-v2} & DIM   & 72.3  & 70.7  & 97.3*  & 72.2  & 32.5  & 30.2  & 20.9  & 56.59 \\
         & TIM   & 62.9  & 57.2  & 98.9*  & 63.3  & 32.9  & 31.8  & 26.4  & 53.34 \\
         & SIM   & 85.8  & 81.8  & \textbf{99.4*} & 82.3  & 61.1  & 54.4  & 46.6  & 73.06 \\
         & Admix & 85.9  & 82.0    & 99.3*  & 82.1  & 61.6  & 52.5  & 45.6  & 72.71 \\
         & S$^2$IM   & 90.1  & 88.6  & 98.1*  & 85.8  & 67.7  & 63.3  & 55.9  & 78.50 \\
         & \cellcolor[rgb]{0.9, 0.9, 0.9} \textbf{STM}   & \cellcolor[rgb]{0.9, 0.9, 0.9} \textbf{91.8} & \cellcolor[rgb]{0.9, 0.9, 0.9} \textbf{91.3} & \cellcolor[rgb]{0.9, 0.9, 0.9} 98.5*  & \cellcolor[rgb]{0.9, 0.9, 0.9} \textbf{87.6} & \cellcolor[rgb]{0.9, 0.9, 0.9} \textbf{76.3} & \cellcolor[rgb]{0.9, 0.9, 0.9} \textbf{71.5} & \cellcolor[rgb]{0.9, 0.9, 0.9} \textbf{64.5} & \cellcolor[rgb]{0.9, 0.9, 0.9} \textbf{83.07} \\
         \hline
         \multirow{6}[2]{*}{Res-101} & DIM   & 78.1  & 75.8  & 67.2  & 99.8*   & 30.8  & 28.6  & 17.8  & 56.90 \\
         & TIM   & 61.3  & 53.6  & 44.1  & 99.5*   & 30.3  & 31.9  & 22.5  & 49.10 \\
         & SIM   & 70.9  & 60.1  & 53.8  & 99.7*   & 26.7  & 26.2  & 15.9  & 50.51 \\
         & Admix & 72.3  & 65.1  & 58.1  & 99.6*   & 26.4  & 27.1  & 16.3  & 52.19 \\
         & S$^2$IM   & 88.4  & \textbf{85.7} & 80.2  & 99.7*   & 49.6  & 46.0    & 33.5  & 69.06 \\
         & \cellcolor[rgb]{0.9, 0.9, 0.9} \textbf{STM}   & \cellcolor[rgb]{0.9, 0.9, 0.9} \textbf{89.3} & \cellcolor[rgb]{0.9, 0.9, 0.9} 85.5  & \cellcolor[rgb]{0.9, 0.9, 0.9}  \textbf{80.8} & \cellcolor[rgb]{0.9, 0.9, 0.9} \textbf{99.9*} & \cellcolor[rgb]{0.9, 0.9, 0.9} \textbf{56.5} & \cellcolor[rgb]{0.9, 0.9, 0.9} \textbf{56.1} & \cellcolor[rgb]{0.9, 0.9, 0.9} \textbf{36.9} & \cellcolor[rgb]{0.9, 0.9, 0.9} \textbf{72.16} \\
         \hline
         \end{tabular}%
      \label{table:1}
  \end{table*}
  
\subsection{Differences with Other Methods}
\label{sec:difference}
\begin{itemize}[leftmargin=*,noitemsep,topsep=2pt]
	\item As shown in Figure \ref{fig:1}, compared with DIM, TIM, SIM, Admix, and S$^2$IM, our STM method introduces generated data from different domains for enhancing the transferability of adversarial examples, while these existing methods mainly adopt data in the same domain for augmentation.
	\item Our STM preserves some original information by mixing the content of original images, while the Admix method obtains diversity of images by mixing images with different categories. 
	\item S$^2$IM transforms the spatial domain into the frequency domain for enhancement, while STM introduces images from different domains based on statistical differences in the low-level features of the dataset.
\end{itemize}

\section{Experiments}
\label{sec:experiment}
In this section, we conduct extensive experiments on the ImageNet-compatible dataset. We first provide the experimental setup. Then we compare the results of the proposed methods with existing methods on both normally trained models and adversarially trained models. Finally, we conduct ablation studies to study the effectiveness of key parameters in our STM. The experimental results were performed multiple times and averaged to ensure the experimental results are reliable.

\subsection{Experimental Settings}
\label{sec:expsetting}
\textbf{Dataset.} We adopt the ImageNet-compatible dataset for our experiments, which is widely used in other works \cite{YuyangLong2022FrequencyDM, JunyoungByun2022ImprovingTT}. It contains 1,000 images with size of $299\times299\times3$, ground-truth labels, and target labels for targeted attacks.

\textbf{Models.} To validate the effectiveness of our methods, we test the attack performance on several popular pre-trained models, \textit{i.e.}, Inception-v3 (Inc-v3) \cite{ChristianSzegedy2016RethinkingTI}, ResNet-50 (Res-50), ResNet-152 (Res-152), Resnet-v2-101 (Res-101) \cite{KaimingHe2015DeepRL}, Inception-v4 (Inc-v4) and Inception-Resnet-v2 (IncRes-v2) \cite{ChristianSzegedy2016Inceptionv4IA}. We also consider adversarially trained models \textit{i.e.}, Inc-v3$_{ens3}$, Inc-v3$_{ens4}$ and IncRes-v2$_{ens}$ \cite{FlorianTramr2017EnsembleAT}.

\textbf{Baselines.} We take five popular input transformation-based state-of-the-art attacks as our baselines, \textit{i.e.}, DIM \cite{YinpengDong2019EvadingDT}, TIM \cite{CihangXie2018ImprovingTO}, SIM \cite{JiadongLin2019NesterovAG}, Admix \cite{XiaosenWang2021AdmixET}, S$^2$IM \cite{YuyangLong2022FrequencyDM}. All these methods are integrated into MI-FGSM~\cite{YinpengDong2017BoostingAA}.

\textbf{Hyper-parameters.} In this work, we set the maximum perturbation $\epsilon = 16$, the number of iterations $T=10$, step size $\alpha=1.6$ and decay factor $\mu=1.0$. We set the transformation probability $p=0.5$ in DIM and the kernel length $k = 7$ in TIM. For SIM and Admix, we use the number of copies $m_1 = 5$, the number of mixed samples $m_2 = 3$, and the admix ratio $\eta=0.2$. For S$^2$IM, we adopt the tuning factor $\rho=0.5$, the standard deviation $\sigma=16$ of $\xi$, and the number of spectrum transformations $N = 20$. For our proposed STM, we set the mixing ratio $\gamma=0.5$, the noise upper bound $\beta=2.0$, and the number of style transfer images $N=20$. 

\subsection{Attack a Single Model}
\label{sec:singleattack}
To validate the effectiveness of our STM, we first compare STM with various input transformation-based attacks, including DIM, TIM, SIM, Admix, and S$^2$IM. All these methods are integrated with the MI-FGSM \cite{YinpengDong2017BoostingAA}. The adversarial examples are generated on Inc-v3, Inc-v4, IncRes-v2, and Res-101, respectively. We report the attack success rates, \textit{i.e.,} the misclassification rates on the crafted adversarial examples in Table~\ref{table:1}.

We observe that STM can effectively improve the attack success rate on black-box models. For example, DIM, Admix, and S$^2$IM achieve the attack success rates of $72.4\%$, $78.6\%$ and $87.5\%$, respectively on Inc-v4 when generating adversarial examples on Inc-v3. In contrast, STM can achieve the attack success rate of $90.8\%$, which outperforms S$^2$IM by a margin of $3.3\%$. On the adversarially trained models, STM consistently exhibits better performance than other input transformation-based methods and improves the average attack success rate by at least $7.45\%$ than other methods. This confirms our motivation that introducing data from different distribution domains for augmentation can enhance the transferability of the adversarial attack, especially for adversarially trained models.

\begin{table}
  \centering
  \caption{Untargeted attack success rates (\%) of various input transformation-based attacks on black-box models in the multi-model setting. The adversarial examples are generated on the ensemble models, \textit{i.e.} Inc-v3, Inc-v4, and IncRes-v2.}
  \vspace{-2mm}
  \setlength{\tabcolsep}{1.5pt}
    \begin{tabular}{|c|ccccc|c|}
    \hline
    Attack & Res-101 & Res-152 & Inc-v3$_{ens3}$ & Inc-v3$_{ens4}$ & IncRes-v2$_{ens}$ & Avg.\\
    \hline
    \hline
    DIM    & 86.5  & 85.4  & 54.4  & 53.4  & 35.8  & 63.08 \\
    TIM    & 75.8  & 73.2  & 56.1  & 52.8  & 42.8  & 60.14 \\
    SIM    & 93.1  & 92.6  & 78.4  & 74.1  & 60.9  & 79.82 \\
    Admix   & 91.8  & 92.4  & 77.6  & 72.4  & 59.3  & 78.80 \\
    S$^2$IM   & 92.7  & 94.3  & 84.2  & 82.9  & 72.8  & 85.38 \\
    \cellcolor[rgb]{0.9, 0.9, 0.9} \textbf{STM}  & \cellcolor[rgb]{0.9, 0.9, 0.9} \textbf{94.8} & \cellcolor[rgb]{0.9, 0.9, 0.9} \textbf{95.6} & \cellcolor[rgb]{0.9, 0.9, 0.9} \textbf{89.4} & \cellcolor[rgb]{0.9, 0.9, 0.9} \textbf{89.1} & \cellcolor[rgb]{0.9, 0.9, 0.9} \textbf{82.6} & \cellcolor[rgb]{0.9, 0.9, 0.9} \textbf{90.30} \\
    \hline
    \end{tabular}%
  \label{table:2}%
\end{table}%

\begin{table*}
  \centering
  \caption{Untargeted attack success rates (\%) of S$^2$IM and our STM when integrated with DIM, TIM, and SIM, respectively. The adversarial examples are generated on Inc-v3. * indicates the white-box model.}
  \vspace{-2mm}
  \setlength{\tabcolsep}{12.0pt}
  \begin{tabular}{|c|ccccccc|c|}
    \hline
    Attack & Inc-v3 & Inc-v4 & IncRes-v2 & Res-101 & Inc-v3$_{ens3}$ & Inc-v3$_{ens4}$ & IncRes-v2$_{ens}$ & Avg.\\
    \hline
    \hline
    S$^2$I-DIM & 99.3*  & 92.9  & 91.5  & 91.2  & 69.5  & 67.7  & 47.8  & 79.99 \\
    \cellcolor[rgb]{0.9, 0.9, 0.9} \textbf{ST-DIM}  & \cellcolor[rgb]{0.9, 0.9, 0.9} \textbf{99.9*} & \cellcolor[rgb]{0.9, 0.9, 0.9} \textbf{93.9} & \cellcolor[rgb]{0.9, 0.9, 0.9} \textbf{93.2} & \cellcolor[rgb]{0.9, 0.9, 0.9} \textbf{92.3} & \cellcolor[rgb]{0.9, 0.9, 0.9} \textbf{75.1} & \cellcolor[rgb]{0.9, 0.9, 0.9} \textbf{75.2} & \cellcolor[rgb]{0.9, 0.9, 0.9} \textbf{53.0} & \cellcolor[rgb]{0.9, 0.9, 0.9} \textbf{83.23} \\
    \hline
    S$^2$I-TIM & 99.3*  & 88.7  & \textbf{87.5} & \textbf{82.8} & 74.5  & 74.2  & 59.7  & 80.96 \\
    \cellcolor[rgb]{0.9, 0.9, 0.9} \textbf{ST-TIM}  & \cellcolor[rgb]{0.9, 0.9, 0.9} \textbf{99.9*} & \cellcolor[rgb]{0.9, 0.9, 0.9} \textbf{89.4} & \cellcolor[rgb]{0.9, 0.9, 0.9} 86.1  & \cellcolor[rgb]{0.9, 0.9, 0.9} \textbf{82.8} & \cellcolor[rgb]{0.9, 0.9, 0.9} \textbf{80.9} & \cellcolor[rgb]{0.9, 0.9, 0.9} \textbf{79.7} & \cellcolor[rgb]{0.9, 0.9, 0.9} \textbf{66.1} & \cellcolor[rgb]{0.9, 0.9, 0.9} \textbf{83.56} \\
    \hline
    S$^2$I-SIM & \textbf{99.8*} & 91.3  & 90.9  & 91.2  & 71.2  & 70.5  & 48.6  & 80.50 \\
    \cellcolor[rgb]{0.9, 0.9, 0.9} \textbf{ST-SIM}  & \cellcolor[rgb]{0.9, 0.9, 0.9} \textbf{99.8*} & \cellcolor[rgb]{0.9, 0.9, 0.9} \textbf{93.4} & \cellcolor[rgb]{0.9, 0.9, 0.9} \textbf{92.9} & \cellcolor[rgb]{0.9, 0.9, 0.9} \textbf{92.0} & \cellcolor[rgb]{0.9, 0.9, 0.9} \textbf{78.9} & \cellcolor[rgb]{0.9, 0.9, 0.9} \textbf{76.3} & \cellcolor[rgb]{0.9, 0.9, 0.9} \textbf{57.3} & \cellcolor[rgb]{0.9, 0.9, 0.9} \textbf{84.37} \\
    \hline
    \end{tabular}%
  \label{table:3}%
\end{table*}%

\begin{table*}[htbp]
  \centering
  \caption{Untargeted attack success rates (\%) on six defense models. The adversarial examples are crafted on the ensemble models, \textit{i.e.,} Inc-v3, Inc-v4 and IncRes-v2.}
    \vspace{-2mm}
    \setlength{\tabcolsep}{13.5pt}
    \begin{tabular}{|c|cccccccc|c|}
    \hline
    Attack & HGD   & R\&P  & NIPs-r3 & Bit-Red & JPEG  & ComDefend & RS & NPR & AVG. \\
    \hline
    \hline
    Admix & 64.0  & 58.1  & 67.8  & 46.7  & 81.7  & 82.9 & 42.3 & 49.1 & 61.6 \\
    S$^2$IM  & 74.2  & 74.2  & 81.0  & 58.6  & 88.4  & 88.7 & 55.2 & 58.9& 72.4 \\
    \cellcolor[rgb]{0.9, 0.9, 0.9} \textbf{STM} & \cellcolor[rgb]{0.9, 0.9, 0.9} \textbf{80.7} & \cellcolor[rgb]{0.9, 0.9, 0.9} \textbf{82.5} & \cellcolor[rgb]{0.9, 0.9, 0.9} \textbf{87.6} & \cellcolor[rgb]{0.9, 0.9, 0.9} \textbf{72.1} & \cellcolor[rgb]{0.9, 0.9, 0.9} \textbf{91.5} & \cellcolor[rgb]{0.9, 0.9, 0.9} \textbf{93.7} & \cellcolor[rgb]{0.9, 0.9, 0.9} \textbf{70.6} & \cellcolor[rgb]{0.9, 0.9, 0.9} \textbf{77.6} & \cellcolor[rgb]{0.9, 0.9, 0.9} \textbf{82.0} \\
    \hline
    \end{tabular}%
  \label{table:4}%
\end{table*}%

\subsection{Attack an Ensemble of Models}
\label{sec:ensattack}
Liu \textit{et al.} \cite{YanpeiLiu2016DelvingIT} have shown that attacking multiple models simultaneously can improve the transferability of the generated adversarial examples. To further demonstrate the efficacy of our proposed STM, we used the ensemble model attack in \cite{YinpengDong2017BoostingAA}, which fuses the logit outputs of various models. The adversaries are generated by integrating three normally trained models, including Inc-v3, Inc-v4, and IncRes-v2. All the ensemble models are assigned equal weights and we test the performance of transferability on two normally trained models and three adversarially trained models.

As shown in Table \ref{table:2}, our proposed STM always achieves the highest attack success rates in the black-box setting. Compared with previous input transformation-based attack methods, STM achieves an average success rate of $90.3\%$ on five black-box models, which outperforms S$^2$IM by an average of $4.92\%$. We also notice that our method has a success rate of over $80\%$ for attacks on all the adversarially training models. This also validates that our method combined with the ensemble model can obtain adversarial examples with higher transferability.


\subsection{Combined with Input Transformation-based Attacks}
\label{sec:comattack}
Existing input transformation-based attacks have shown great compatibility with each other. Similarly, our method can also be combined with other input transformation-based methods to improve the transferability of adversarial examples. To further demonstrate the efficacy of the proposed STM, we compare the attack success rates of S$^2$IM (known as the best method) and our STM when combined with DIM, TIM, and SIM, respectively. We generate adversarial examples on the Inc-v3 model and test the transferability of adversarial examples on six black-box models. 

As shown in Table \ref{table:3}, under the same setting, STM performs much better than S$^2$IM when combined with various input transformation-based attacks. On average, STM outperforms S$^2$IM by a clear margin of $3.24\%$, $2.6\%$ and $3.87\%$ when combined with DIM, TIM and SIM, respectively. Especially, our STM tends to achieve much better results on the ensemble adversarially trained models, which have shown great effectiveness in blocking the transferable adversarial examples. Such consistent and remarkable improvement supports its high compatibility with existing input transformation-based attacks and further validates its superirority in boosting adversarial transferability.

\subsection{Attack Defense Models}
\label{sec:defense}
In this subsection, besides normally trained models and adversarially trained models, we further validate the effectiveness of our methods on other defenses, including Bit-Red \cite{xu2018feature}, ComDefend \cite{jia2019comdefend}, JPEG \cite{guo2018countering}, HGD \cite{liao2018defense}, R\&P \cite{xie2018mitigating}, NIPS-r3 \cite{naseer2020self}, RS \cite{cohen2019certified} and NPR \cite{naseer2020self}. The adversarial examples are generated in the same setting as in Section~\ref{sec:ensattack}.
More specifically, adversarial examples are generated on an ensemble of Inc-v3, Inc-v4, and IncRes-v2, and the weight for each model is $1/3$. 

The experimental results are shown in Table \ref{table:4}. In the setting of ensemble models, we can observe that our algorithm can significantly boost existing attacks. For example, Admix and S$^2$IM only attain an average success rate of $66.9\%$ and $77.5\%$ on the six defense models, respectively, while our STM can achieve an average rate of $84.7\%$, which is $17.8\%$ and $7.2\%$ higher than them, respectively. This demonstrates the remarkable effectiveness of our proposed method against both adversarially trained models and other defense models and brings a greater threat to advanced defense models.

\begin{table*}[h]
  \centering
  \caption{Classification accuracy (\%) of the ImageNet-compatible dataset on original images and the transformed images by different strategies, such as image style transformation, fine-tuning the model, and mixing up the original image content.}
  \setlength{\tabcolsep}{9.30pt}
    \begin{tabular}{|ccc|cccccc|c|}
    \hline
    Style transfer & Fine-tuning & Mixing up & Inc-v3 & Inc-v4 & IncRes-v2 & Res-50 & Res-101 & Res-152 & AVG.\\
    \hline
    \hline
     \ding{55} & \ding{55} & \ding{55} & 95.1 & 97.6 & 100.0 & 83.3 & 85.4 & 87.3 & 91.45\\
     \hline
     \ding{51} & \ding{55} & \ding{55} & 38.2  & 47.2  & 49.7  & 11.6  & 17.8  & 18.6 & 30.52\\
     \ding{51} & \ding{51} & \ding{55} & 61.0    & 67.0    & 72.5  & 27.7  & 37.6  & 37.5 & 50.55\\
     \ding{51} & \ding{51} & \ding{51} & \textbf{88.2}  & \textbf{92.2}  & \textbf{95.1}  & \textbf{68.1}  & \textbf{70.2}  & \textbf{75.6} & \textbf{81.57}\\
    \hline
    \end{tabular}%
  \label{table:5}%
\end{table*}%

\begin{table*}[h]
	\centering
	\caption{Untargeted attack success rates (\%) on black-box models when applying different strategies, such as image style transformation, fine-tuning the model, and mixing up the original image content. Each strategy adds random noise by default. The adversarial examples are crafted on Inc-v3.}
	\vspace{-2mm}
	\setlength{\tabcolsep}{8.80pt}
	\begin{tabular}{|ccc|cccccc|}
		\hline
		Style transfer & Fine-tuning & Mixing up & Inc-v4 & IncRes-v2 & Res-101 & Inc-v3$_{ens3}$ & Inc-v3$_{ens4}$ & IncRes-v2$_{ens}$  \\
		\hline
		\hline
		\ding{55} & \ding{55} & \ding{55} & 49.7 & 47.1 & 61.9 & 22.3  & 23.4 & 10.9 \\
            \hline
		\ding{51} & \ding{55} & \ding{55} & 57.8 & 55.3 & 64.7  & 36.5 & 34.4 & 18.8  \\
		\ding{51} & \ding{51} & \ding{55} & 71.9 &70.3 & 72.3  & 46.4  & 45.3  &26.0\\
		\ding{51} & \ding{51} & \ding{51} & \textbf{90.8} & \textbf{90.1} & \textbf{82.4} & \textbf{68.3} & \textbf{68.1} & \textbf{46.3} \\
		\hline
	\end{tabular}%
	\label{table:6}%
\end{table*}%


\subsection{Ablation Studies}
\label{sec: ablation}
In this subsection, we conduct a series of ablation experiments to study the impact of fine-tuning the style transfer network and mixing the original image. To simplify the analysis, we only consider the transferability of adversarial examples crafted on Inc-v3 by the STM method.

Since the stylized images may mislead the surrogate model, this will lead to imprecise gradient information during iterations and affect the success rate of adversarial attacks. To address this issue, we first fine-tune the style transfer network so that the generated images can be correctly classified by multiple models. We also mix up the original image with its style-transformed images to further avoid such imprecise gradients.
a) We first test the classification accuracy of the stylized images by adding these two strategies to the classification network and evaluate the performance of the black-box attacks with these strategies. As shown in Table \ref{table:5}, when we transform the original images into stylized images, the prediction accuracy of stylized images decreases. This indicates that stylized images might not maintain semantic consistency for classification networks. The results also show that fine-tuning the model and mixing up the original image content can effectively maintain the semantic labels of the stylized images. We can further observe that mixing up the content of the original image can significantly maintain semantic consistency, which avoids generating imprecise gradient information during the iterative optimization process. b) As shown in Table \ref{table:6}, we find that all of these strategies are beneficial to improve the transferability of adversarial examples, especially for mixing up the original image content. And fine-tuning the style transfer network can significantly improve the transferability, validating our assumption that maintaining semantic consistency can boost adversarial transferability. When combining all of these strategies, STM achieves the best performance, supporting our rational design.

\section{Conclusion}
\label{sec:conclu}
Inspired by the fact that the \textit{domain bias} issue affects the generalization ability of the normally trained models, we postulate that it might also impact the transferability of the adversarial examples. However, we find that existing input transformation-based methods mainly adopt the transformed data in the same source domain, which might limit the adversarial transferability. Based on this finding, we propose a novel attack method named Style Transfer Method (STM), which transforms the data into different domains using an arbitrary style transfer network to enhance the adversarial transferability. To maintain semantic consistency and avoid stylized images misleading the surrogate model and leading to imprecise gradients during iterative process, we fine-tune the style transfer network and mix up the content of the original image with its style-transformed images. Our method can be well integrated with existing input transformation-based methods to further improve adversarial transferability. Empirical results on the ImageNet-compatible dataset demonstrated that our STM can achieve higher attack success rates in both normally trained models and adversarially trained models, with excellent performance on both non-target and targeted attacks than state-of-the-art input transformation-based attacks.

\begin{acks}
This work was supported by the National Natural Science Foundation of China (Nos.\ 62072334, 61976164, and 6227071567), and the National Science Basic Research Plan in Shaanxi Province of China (No.\ 2022GY-061).
\end{acks}

\bibliographystyle{ACM-Reference-Format}
\bibliography{sample-base}

\end{document}